\documentclass[fleqn,10pt]{wlscirep}
\usepackage[utf8]{inputenc}
\usepackage[T1]{fontenc}

\usepackage{cite}
\usepackage{amsmath,amssymb,amsfonts}
\usepackage{algorithmic}
\usepackage{graphicx}
\usepackage{textcomp}
\usepackage{latexsym}
\usepackage{booktabs}
\usepackage{dsfont}
\usepackage{adjustbox}
\usepackage{url}
\usepackage{hyperref}
\hypersetup{
	final,
	breaklinks,
	colorlinks,
	linkcolor={red!69!black},
	citecolor={green!69!black},
	urlcolor={blue!69!black}
}
\usepackage[symbol]{footmisc}

\usepackage[numbers,square]{natbib}
\usepackage[capitalize]{cleveref}
\Crefname{section}{Section}{Sections}
\Crefname{table}{Table}{Tables}
\Crefname{figure}{Figure}{Figures}

\newcommand{\ie}{\textit{i}.\textit{e}.~}
\newcommand{\eg}{\textit{e}.\textit{g}.~}
\newcommand{\etc}{\textit{etc}.}
\newcommand{\rv}[1]{\textcolor{black}{#1}}

\title{Show from Tell: Audio-Visual Modelling\\ in Clinical Settings}

\author[,1,2]{Jianbo Jiao\footnote[2]{These authors contributed equally to this work}\footnote[1]{Corresponding authors: {\{jianbo.jiao, alison.noble\}@eng.ox.ac.uk}}}
\author[$\dagger$,1,3]{Mohammad Alsharid}
\author[4,5]{Lior Drukker}
\author[4]{Aris T. Papageorghiou}
\author[1]{\\Andrew Zisserman}
\author[*,1]{J. Alison Noble}
\affil[1]{Department of Engineering Science, University of Oxford, Oxford, UK}
\affil[2]{School of Computer Science, University of Birmingham, Birmingham, UK}
\affil[3]{Department of Electrical Engineering and Computer Science, Khalifa University, Abu Dhabi, United Arab Emirates}
\affil[4]{Nuffield Department of Women's \& Reproductive Health, University of Oxford, Oxford, UK}
\affil[5]{Rabin Medical Center, Tel-Aviv University Faculty of Medicine, Israel}

\begin{abstract}
Auditory and visual signals usually present together and correlate with each other, not only in natural environments but also in clinical settings. However, the audio-visual modelling in the latter case can be more challenging, due to the different sources of audio/video signals and the noise (both signal-level and semantic-level) in auditory signals -- usually speech.
In this paper, we consider audio-visual modelling in a clinical setting, providing a solution to learn medical representations that benefit various clinical tasks, without human expert annotation. A simple yet effective multi-modal self-supervised learning framework is proposed for this purpose. The proposed approach is able to localise anatomical regions of interest during ultrasound imaging, with only speech audio as a reference.
Experimental evaluations on a large-scale clinical multi-modal ultrasound video dataset show that the proposed self-supervised method learns good transferable anatomical representations that boost the performance of automated downstream clinical tasks, even outperforming fully-supervised solutions.
\end{abstract}
\begin{document}

\flushbottom
\maketitle

\thispagestyle{empty}

\section*{Introduction}
\label{sec:intro}

Visual signals are often accompanied by auditory signals which provide essential assistance for scene understanding. 
Such multi-modal scenarios widely exist in both natural environments and clinical settings.
Our goal is to be able to learn visual capabilities (such as localising \emph{kidneys}, or detecting standard planes in an ultrasound scan) in clinical scenarios simply by listening to sonographers (\ie operators) describing what they are doing.

Audio-visual modelling has attracted interest in the recent literature~\cite{arandjelovic2017look,arandjelovic2018objects,kazakos2019TBN}, in which the basic assumption is that the underlying video and audio signals are densely correlated with each other. Based on this assumption, correspondence modelling  is leveraged, \ie examining if the input audio and video signals correspond or not. Such a binary discrimination model is simple, but has been shown to be effective in learning audio-visual representations.
Some work~\cite{korbar2018cooperative,morgado2021audio} has proposed contrastive learning for audio-visual modelling by referring to negative samples. This helps model discrimination ability and has been shown to result in a better representation quality~\cite{morgado2021audio}.

\begin{figure*}
\includegraphics[width=\textwidth]{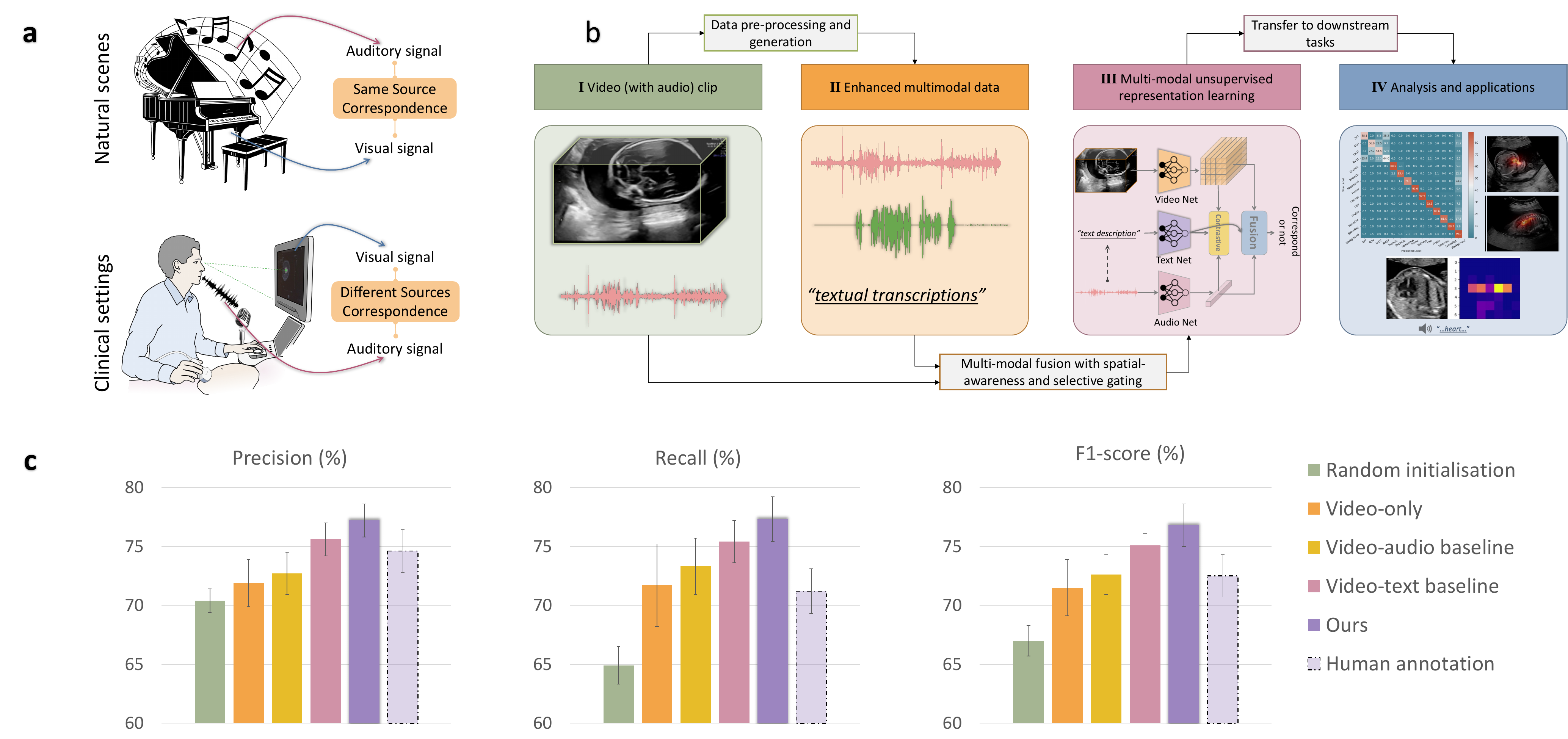}
\caption{\rv{\textbf{Multi-modal representation learning in clinical setting.} (\textbf{a}) Illustration of audio-visual modelling in a natural scene setting (top) and in a clinical setting (bottom). (\textbf{b}) Pipeline for audio-video modelling and analysis in clinical settings from raw video footage. Video frames with the corresponding speech audio are extracted from the raw footage (I). After data pre-processing and text data generation, the enhanced multi-modal data (II) are fed into a joint fusion framework to learn multi-modal representations without human annotation (III). The whole automated system can be transferred to several downstream tasks and used for large-scale analysis (IV). (\textbf{c}) Evaluation results on the downstream task of standard plane detection. Note the method with dotted borders is fully-supervised using external human annotations.}}
\label{fig:fig1}
\end{figure*}

Promising progress has been made in the above-mentioned literature for audio-visual modelling in natural scenes. However,  in clinical scenarios the setting is different and consequently it is more challenging to model (see Figure~\ref{fig:fig1}a).
Different from the natural scenario, in a clinical setting the auditory signal and visual signal usually arise from different sources, and have sparse correlation, \eg the conversational speech audio and the corresponding captured medical images/videos. This prevents the direct application of existing methods that were designed for natural scenes.
Similar to natural scenes, the modelling of audio-visual signals is helpful - in this case for learning medical representations without human supervision, especially expert manual annotations, which are costly and hard to scale. During real-time ultrasound (US) imaging narrations from clinicians are describing the visual signals and as a result they have underlying strong correspondence with the anatomy seen on screen. Our hypothesis is that modelling such correlations in a self-supervised manner will be beneficial for automating downstream clinical tasks.

To this end, in this paper, we consider audio-visual modelling in a clinical setting and propose a simple framework (Figure~\ref{fig:fig1}b) to address the challenges faced in doing this.
Here the \emph{clinical setting} we used is fetal US scanning, where audio recordings containing the speech uttered by the sonographers during the scan are considered, together with the US video shown on the screen. We can do this because  sonographers speak about what they see on the screen during an examination, as an explanation to the pregnant individual being scanned.
More specifically, we start with modelling the first-order correspondence between the speech audio and its corresponding video signal. Cross-modal contrastive learning is then introduced to strengthen the discrimination between the current sample and unrelated negative samples. Clinical audio recordings are noisy.  Therefore, we propose to pre-process the audio data to enhance the related speech content before feeding it to the deep model. To enrich the semantic understanding of the audio data and bootstrap the audio-visual modelling, a text branch is introduced to the proposed framework, forming a multi-modal modelling framework. The text is automatically transcribed from the audio signal, and we propose a learnable gating module towards the end of this text branch to spot the text embeddings of the relevant anatomies.
\rv{Note that in our clinical ultrasound setting, the originally captured data are the fetal ultrasound video from the ultrasound machine and the speech audio from the sonographers. The text information is not available and accurate transcription is infeasible to achieve. Thus our work aims at audio-visual modelling and takes textual data as intermediate information to help the modelling.}

In addition to learning self-supervised cross-modal representations, we demonstrate audio-guided visual localisation by directly leveraging the learned representations. This could be used in various clinical applications, \eg better user experience for patients during scanning, training new sonographers and low-cost data annotation for region-of-interest by only using audio, to name a few.
Note that during the inference stage (\ie in real applications), no text information is needed and only the audio signal is sufficient to help the visual localisation.
We perform extensive experimental evaluations on a large-scale multi-modal clinical US dataset to validate the effectiveness of the proposed approach by transferring the learned representations to downstream tasks. Analysis of three downstream clinical tasks shows that the proposed approach learns good transferable representations, which surpass the performance of fully-supervised solutions.
To summarise, the main contributions of this work are as follows:
\begin{itemize}
\item We propose, to our knowledge, the first audio-visual modelling framework with generated text as an additional modality for representation learning, in a clinical setting where the correlation modelling for audio-visual data is more challenging than natural scenes.
\item We introduce several strategies to address the challenges in clinical settings, including audio pre-processing, selective gating module for text embedding refinement and multi-modal joint optimisation.
\item Experimental analysis validates the effectiveness of the proposed framework, with better performance than alternative solutions and superior to fully-supervised settings. We also present an approach to visually localise anatomical regions of interest according to audio guidance.
\end{itemize}

\section*{Methods}

\begin{figure*}[t]
\includegraphics[width=\textwidth]{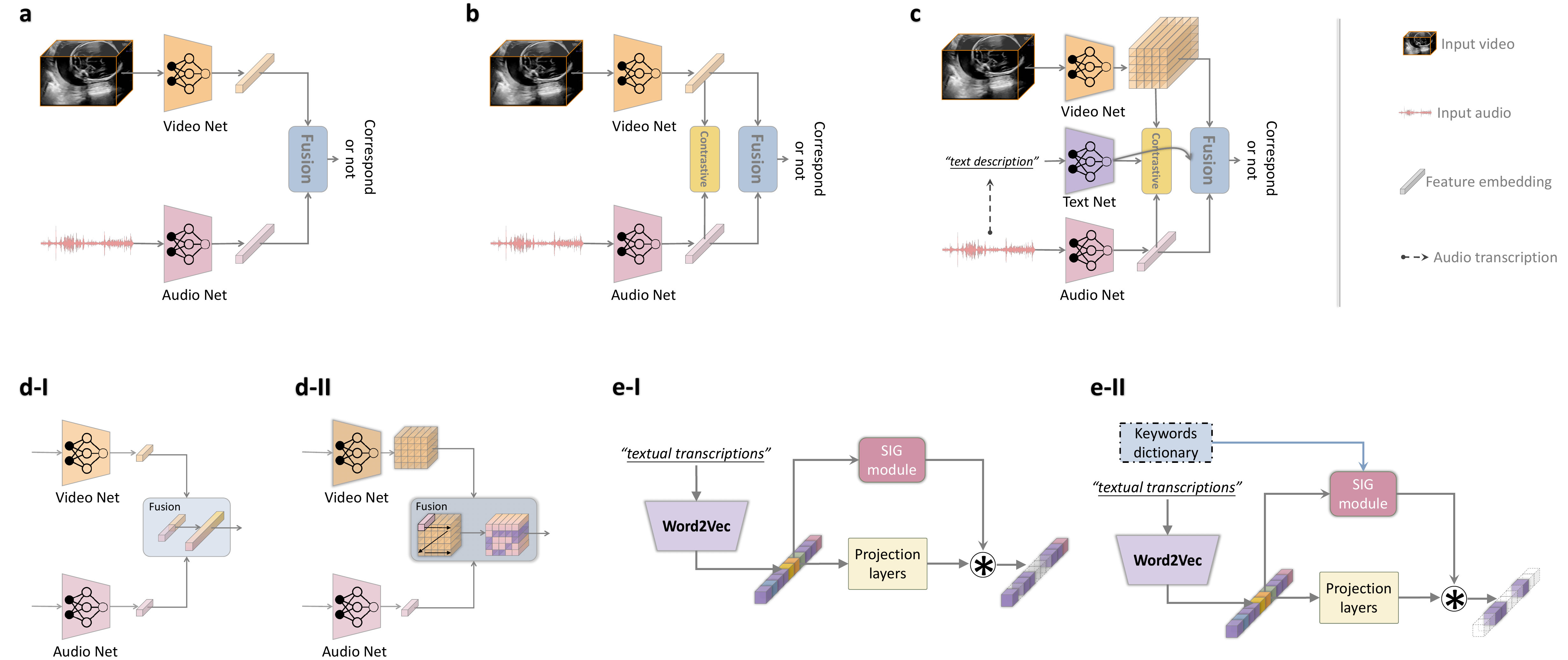}
\caption{\textbf{Audio-video modelling architectures.} (\textbf{a}) Baseline model of correspondence modelling. (\textbf{b}) Correspondence modelling with contrastive learning. (\textbf{c}) The proposed multi-modal modelling approach with additional textual information and visual-spatial awareness.
(\textbf{d-I}) Basic fusion strategy by concatenating multi-modal features. (\textbf{d-II}) Spatial-aware fusion by element-wise product.
(\textbf{e-I}) The proposed selective information gating (SIG) module, with the coarse level of only filtering outliers. (\textbf{e-II}) The SIG module with fine-grained level of keywords spotting by only focusing on specific keywords of interest. The $\circledast$ means element-wise product.}
\label{fig:arch}
\end{figure*}

We propose a framework to model the audio-visual representations by leveraging intermediate text information. The proposed approach has three original components: audio concentration for audio data processing, a selective gating module for embedding refinement, and a multi-modal correlation optimisation loss.
Figure~\ref{fig:arch} illustrates the framework of audio-video modelling with the main ideas of the proposed method shown in Figure~\ref{fig:arch}c.

\subsection*{Large-Scale Multi-Modal Clinical Dataset}
The proposed approach is validated on a large-scale multi-modal clinical US dataset called the PULSE (Perception Ultrasound by Learning Sonographer Experience) dataset which is described in~\cite{drukker2021transforming}. Ethics approval was granted by the West of Scotland Research Ethics Service, UK Research Ethics Committee (Reference 18/WS/0051).
Informed consent was obtained from all participants, and all methods were performed in accordance with the relevant guidelines and regulations.
Briefly, the PULSE dataset consists of full-length US video recorded with concurrent speech audio, eye-gaze tracking data and probe motion data, as sonographers perform routine scans on pregnant women. A typical scan setting is shown in Figure~\ref{fig:fig1}a bottom.
In this work, we focus on modelling the video and audio data for the second-trimester data (\ie the 20-week anomaly scan).
The video data was frame-grabbed from the US machine in real-time to a computer in a fully anonymised fashion. The original full-length video was stored at 30 FPS using lossless compression. The speech audio data was recorded in synchronisation with the video data by using microphones placed in the scanning room.

\subsection*{Multi-Modal Medical Representation Learning}
\label{sec:baseline}
\paragraph{Baselines}
We first present a baseline to model the correlation between the US video and audio data modalities. Specifically, the most intuitive approach is to predict if the given video and audio samples correspond or not, as usually done in natural scenes. To this end, we use a binary classification framework to model audio-visual correspondence. As illustrated in Figure~\ref{fig:arch}a, the US video clip and audio segment are input to the video and audio sub-networks to obtain corresponding feature embeddings, followed by a fusion operation to produce the final feature for correspondence prediction. The original recorded video with the accompanying synchronised audio signals are assumed to correspond with each other, while shifted video-audio data pairs or pairs from different scans are considered to be non-corresponding.
Let the video signal be $V$ and the audio signal $A$. The correspondence modelling task can be simply modelled as a binary classification problem:
\begin{equation}\label{eq:base}
	L_{base}(V_{i},A_{i})=-\frac{1}{N} \sum_{i=1}^{N} c_{i} \cdot \log \left(p\left(A_i,V_i\right)\right)+\left(1-c_i\right) \cdot \log \left(1-p\left(A_i,V_i\right)\right),
\end{equation}
where $c$ is the correspondence label (0 or 1) and $p$ is the predicted probability of the audio-video pair being in correspondence. $N$ is the number of samples.
\rv{Note here the label $c$ does not need any human annotations, but instead naturally encoded in the data. Specifically, when the audio and video signals are aligned with each other (by default naturally as they are in the dataset), $c=1$, otherwise, $c=0$ if they have not been matched with the signal that they originally correspond to in the dataset.}

\paragraph{Cross-modal contrastive learning}
The above baseline model only describes the alignment of the audio and video data, being a very strict constraint that may not be sufficient to model the relationship between these two modalities due to the challenges in the clinical setting.
\rv{We observe that in our speech audio data, there are some medical-unrelated contents (\eg random talk), which deteriorate the correlation learning between the two modalities.}
Contrastive learning aims to pull the similar data pairs closer while pushing dissimilar ones away from each other, in the embedding space during optimisation. In addition to modelling the relationship of the positive samples that are aligned at the same timestamp, it also considers dissimilar sample pairs, which act as negative samples for the modelling.
As a result, built upon the baseline model, cross-modal contrastive learning is introduced to model the audio-video representation\rv{, and encourage a soft constraint towards the audio-visual relationship modelling}. Specifically, consider a positive pair of audio-video data as $(A_i, V_i)$, and other sample pairs $(A_i, V_j)$ that are misaligned (\ie $i\neq j$) as the negative pairs. Then the loss function for contrastive learning $L_{contra}$ is defined by:
\begin{equation}\label{eq:contra}
L_{contra}(V_{i},A_{i})=-\log \frac{\operatorname{sim}\left(V_{i}, A_{i}\right)}{\sum_{j=1}^{\rv{N}} \mathds{1}_{[j \neq i]} \operatorname{sim}\left(V_{j}, A_{i}\right)},
\end{equation}
where $\mathds{1}_{[j \neq i]}$ equals 1 iff $j \neq i$ and otherwise 0. The sim() function is a similarity measurement over the exponential of the arguments.
\rv{Though intra-modal contrastive learning may be able to boost the representation learning ability if further included, here we focus on cross-modal contrastive learning. And we found in a recent work~\cite{eccvw22} that it may lead to some issues in clinical settings, where additional anatomy annotations may be required. As a result, in this paper we focus on the cross-modal learning approach.}

\paragraph{Spatial awareness}
Although contrastive learning provides an adaptive correlation modelling solution, the conventional similarity measure, \ie a one-dimensional feature vector (as shown in Figures~\ref{fig:arch}a and \ref{fig:arch}b), ignores spatial context in the feature embedding and loses local spatial awareness. 
To address this, instead of compressing the output bottleneck feature of the video network to be a 1-D vector, we preserve the spatial dimensions of the feature before fusing it with the audio feature (Figure~\ref{fig:arch}c). 
\rv{Specifically, the spatial dimensions are achieved by keeping the spatial feature map from the second last layer in the video network instead of the 1-D feature vector after the last pooling layer as in the above settings.}
In this way, the merged features would account for not only global information, but also local structural patterns (anatomies). Here, the audio-video fusion is achieved by element-wise \rv{dot product between the audio feature vector and each of the visual feature vector in the visual feature map, 
followed by global average pooling and fully-connected layers in an attention-like aggregation manner into a single vector before being fed into the loss functions. The element-wise dot product is used to measure the similarity between the audio feature and the visual features, aiming to activate the high-response regions on the spatial visual feature map.} 
As illustrated in Figure~\ref{fig:arch}d, by using the spatial-aware fusion strategy, the multi-modal features are able to capture local spatial information instead of global semantics only (as in Figure~\ref{fig:arch}d-I).

In this case, the final fused feature is a three-dimensional tensor (instead of one-dimensional vector) with spatial information that represents the spatial relationship between the visual and audio signals.
The corresponding loss function used for training is then defined as:
\begin{equation}
L_{base}(V_i^s,A_i) + L_{contra}(V_i^s,A_i),
\end{equation}
where $V_i^s$ is the 3-D visual feature with spatial information.

\paragraph{Adding textual information}
The speech audio data modality provides auxiliary supportive information to assist in video understanding and representation learning, but according to our observations, in our application, it can also bring ambiguity in correlation modelling. Specifically, the audio signal is not generated from the visual source and as a result, may have a weak (or even no) correlation with video content. For example, the content of the audio is often a conversation between the sonographer and the subject while the video being shown on the screen is that of fetal US content. There are even different kinds of background sounds that can exist in the audio data (we have observed ambulance sirens, people coming into the scanning room, etc). These challenges prevent the model from learning accurate and robust multi-modal representations.
To this end, we propose to leverage additional information along with the audio signal, by including a third data modality -- text.
We automatically labelled the speech data using automatic speech recognition (ASR). Textual transcriptions are acquired by applying an ASR model to the speech audio data. In this way, we can use the textual feature embedding to help the multi-modal US representation learning, and this also allows us to focus on key information (\eg anatomy-related terminology) while eliminating unrelated information by keyword spotting.
The basic idea is illustrated in Figure~\ref{fig:arch}c. Specifically, we use the word2vec model~\cite{mikolov2013distributed} followed by two fully-connected layers to project textual information into word embeddings.

However, due to the poor performance of the ASR model when applying to unseen US data, the ASR transcribed text cannot be directly leveraged as input for the model training. Apart from the existing challenges in audio, the ASR transcription has many errors (\eg non-meaningful English), which may further confuse the model during the joint representation learning.
Therefore, we propose approaches to address these issues (\ie noise in audio, ambiguity in transcribed text, \etc), as will be elaborated on in the following sections.

\subsection*{Selective Information Gating}
As mentioned above, the transcribed raw textual data is inappropriate to be directly fed into the text embedding model. In this section, we introduce a new selective information gating (SIG) module to filter out the undesired information and only preserve the key information that we are interested in in a selective manner.
The proposed SIG module is designed in two variants: outlier filtering and keywords spotting, as illustrated in Figure~\ref{fig:arch}e and detailed below.

\paragraph{Filtering outliers}
The first SIG approach is at a relatively coarse level, where only outliers are removed. \rv{Here we consider those wrongly transcribed words \eg~invalid English words and blank text segments (zero padding) as outliers.} Whenever such outliers are detected in the input text, the corresponding embeddings are removed and thus do not contribute to the final correlation modelling. \rv{The wrongly transcribed words are effectively identified before being fed into the SIG module, and the module simply filters them out to remove their possible influence on model training. These words have been identified through the tokenizer used with the word2vec embedding layer, and such `unknown words' that do not exist in the word2vec dictionary would be represented with a $\mathrm{\langle unk\rangle}$ token. Then it is the word2vec embedding of this token that gets filtered out by the SIG module.} More specifically, the SIG module \rv{consists of a convolutional layer with $1\times1$ kernel size followed by a fully-connected layer}, which outputs a weighting vector to combine with the word embeddings, as shown in Figure~\ref{fig:arch}e-I. The projection layers are fully-connected layers to project the word2vec embeddings to combine with the SIG module output. The final filtered text embedding is used to optimise the multi-modal learning objective.

\paragraph{Keywords spotting}
Even when the ASR model is able to correctly recognise all the remaining speech from the SIG output, there can be some content that is not related to the video. For instance, another medical professional enters the room and has a conversation with the sonographer that is unrelated to the US scan. In this case, modelling the text (as well as the audio) with the video would confuse the model. To reduce this potential source or error, we introduce a \emph{keywords dictionary} to the SIG module, so that the SIG module is aware of the keywords and select the corresponding information (\cref{fig:arch}e-II). Specifically, the SIG module only focuses on the keywords within the dictionary when predicting the weights for the gating operation\rv{, in a compare and matching manner by filtering out and only selecting those words exist in the dictionary while depressing the features of other words}.
\rv{The keywords in the dictionary are constructed empirically with domain knowledge from the clinicians.}

\begin{figure*}
\includegraphics[width=\textwidth]{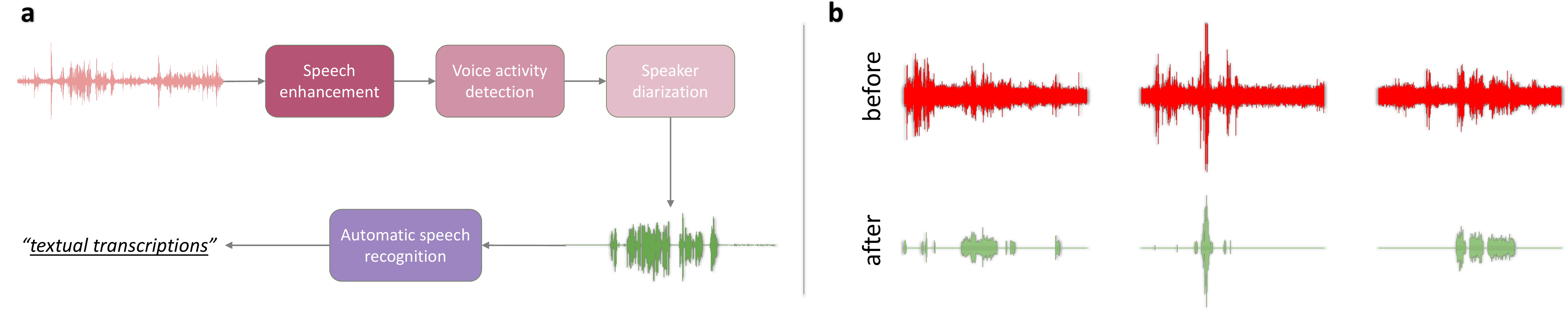}
\caption{\textbf{Audio data cleaning.} (\textbf{a}) Pipeline for the proposed audio data cleaning. The input audio signal (the left red waveform) results in an enhanced audio signal (the bottom-right green waveform) from the sonographer accompanied by a corresponding text segment (the bottom-left text). (\textbf{b}) Example audio data samples \emph{before} (top) and \emph{after} (bottom) the proposed audio data pre-processing.}
\label{fig:audproc}
\end{figure*}

\subsection*{Multi-Modal Optimisation}
To optimise the multi-modal joint representation learning, we propose to use a combined loss function that consists of correspondence modelling and contrastive learning for the video-audio-text modalities.
More specifically, according to the definitions in equation~(\ref{eq:base}) and equation~(\ref{eq:contra}), the final joint loss function for multi-modal optimisation is defined as
\begin{equation}\label{eq:loss}
\mathcal{L}=\alpha\cdot L_{base}(V_i,A_i) + \beta\cdot L_{base}(V_i,T_i) + \gamma\cdot L_{contra}(V_i,A_i) + \delta\cdot L_{contra}(V_i,T_i),
\end{equation}
where $T_i$ is the text sample and $\{\alpha,\beta,\gamma,\delta \mid \alpha+\beta+\gamma+\delta=1\}$ are weighting parameters to balance the contribution of each loss term.

\section*{Results}

In this section, we first describe the detailed experimental settings including the data and evaluation strategies. We will also compare with other alternative or state-of-the-art solutions, with isolated ablation studies to validate the effectiveness of the proposed approach.


\subsection*{Data}\label{sec:data}
Full-length routine fetal second-trimester US videos were acquired by expert sonographers as part of the PULSE study. \rv{As there is no public dataset available with the same clinical setting as ours to our knowledge, here the experiments are performed on our PULSE data.} Sonographer audio recordings that include audible vocal descriptions of the performed scans were collected with the videos, creating an audio-visual dataset. In total, there are 81 scans with such audio and video data and each scan contains around 55,000 frames. Each whole scan was divided into smaller video clips to feed into the networks, resulting in a total of 73,681 clips.
The speech audio data was transcribed via ASR, allowing for the inclusion of the third modality, text. 
As mentioned above, the text was further processed by filtering unrelated words and focusing on the keywords for the text embedding extraction. The keywords dictionary consists of anatomical terminologies within the fetal US scenario.
When sampling the positive and negative training pair, for each video sample at a particular timestamp $t$, the corresponding audio and text are sampled in a neighbour region with the time range of 0.6 s (\ie $t \pm 0.3s$ of the video sample at $t$). Before feeding into the audio network, the original audio data was converted to the 2-dimensional log-spectrogram format with size $256\times256$ using a short-time Fourier transform with frequency bands of 256, window length 10ms and hop length of 5ms.
For the aforementioned downstream tasks, we used an additional set of 135 scans and conducted three-fold cross-validation (by randomly selecting 90 scans for training in each fold).

\subsection*{Audio Data Cleaning}

An audio cleaning step was implemented as illustrated in Figure~\ref{fig:audproc}a.  The raw audio data contains unwanted background signals (including the sound of US machine keyboard actions and the room air conditioner fan among other background sounds) as well as irrelevant, undesired speech. The first step to reduce undesired audio is to apply audio denoising followed by voice activity detection, in which the former is achieved by speech enhancement~\cite{defossez2020real} and the latter by applying VADNet~\cite{wagner2018deep}. That makes it possible to differentiate speech segments from noise segments as well as seconds of silence.

After background noise reduction, we further separate the audio signal into different speakers using speaker diarization. Speaker diarization is a speech processing technique that aims to separate an audio recording into two (or more) speech segments where each segment contains a single unique speaker. Specifically, we employ the pyannote-audio \cite{bredin2020pyannote} framework to perform speaker diarization and assume that the speaker with the most speech segments is the sonographer. By only preserving the sonographer's speech and disposing of the speech of other people, we have a ``clean'' speech dataset suitable for analysis. 
Examples of the audio data before the pre-processing and after the pre-processing are presented in~\cref{fig:audproc}b for reference.

Finally, transcribing the sonographer speech segments with a fine-tuned wav2vec-based automatic speech recognition (ASR) model \cite{baevski2020wav2vec}, we identify the speech segments where anatomically relevant information is being spoken by comparing the transcribed text with an existing dictionary of anatomically relevant keywords. 


\subsection*{Assessment and Evaluation Criteria}\label{sec:ass}
After learning the multi-modal representations, we need an assessment tool to evaluate the quality of the learned representations and their applications. Following prior works~\cite{arandjelovic2017look,jiao2020self}, we use several downstream tasks to assess the quality of these  representations, in a transfer learning manner. Here, in our fetal US scan setting, three downstream tasks are chosen: standard plane detection, eye-gaze saliency prediction, and audio-guided visual localisation. Through these three tasks, we not only evaluate the learned visual representations (as in prior works) but also evaluate the quality of the joint representation (by the third task).
The standard plane detection is a classification problem, where given an US image, the task aims to classify it into an anatomical category. The eye-gaze saliency prediction is a regression task that tries to predict the eye-gaze attention regions that the sonographer is focusing on. The visual localisation task aims to localise the anatomy regions in the US video according to the indication from the corresponding audio data. The first two downstream tasks focus more on the visual representations, while the third one needs the model to have a thorough multi-modal understanding so that it can successfully localise the audio-guided visual anatomies.
For quantitative assessment, we leverage the commonly used \emph{precision, recall} and \emph{F1} score to evaluate the standard plane detection task; the Kullback-Leibler divergence (\emph{KL}), normalised scanpath saliency (\emph{NSS}), area under ROC curve (\emph{AUC}), Pearson's correlation coefficient (\emph{CC}) and similarity (\emph{SIM})~\cite{bylinskii2018different} for the saliency prediction task; \rv{as there is no localisation annotations available in our dataset and it is hard to acquire accurate pixel-level annotations, it is infeasible to present a quantitative evaluation on the third task. Therefore, we consider eye-gaze saliency to be a close enough approximation to expert-made annotations as the eye gaze data comes from the experts themselves. Eye-gaze saliency has been used in other work~\cite{sugano2016seeing, alsharid2022gaze} as well in an effort to tie language with visual regions of interest. We provide a corresponding quantitative evaluation in this case.}

We also found that the saliency prediction in the clinical setting is different from that in natural vision scenarios~\cite{bylinskii2018different}, as we focus on more fine-grained anatomical local structures in medical imaging. As a result, in addition to the aforementioned metrics, we further design a new metric tailored for eye-gaze saliency assessment in the clinical setting. 
This metric is designed to model the compactness of the predicted saliency region, as well as the intensity of the highly-confident response. Specifically, assuming the area of the whole saliency regions to be $\alpha_S=\|I_S\|_0$ ($I_S$ is the normalised saliency map) and the highly-confident response region as $\alpha_H=\|I_{HS}\|_0$ where $I_{HS}=I_S>thres$, the high response intensity is defined as $\eta_H=\alpha_H/\alpha_S$. Additionally, we also consider the correctness of the predicted saliency when measuring its compactness, by using the intersection-over-union (IoU) metric (IoU=intersection($I_S,\hat{I_S}$)/union($I_S,\hat{I_S}$)). The proposed compactness metric is then defined as
\begin{equation}
Comp=\operatorname{IoU}\cdot\frac{\eta_H}{\alpha_S}. 
\end{equation}

\subsection*{Implementation Details}\label{sec:impl}
We follow our preliminary work~\cite{jiao2020selfvs} for the baseline network design and the parameter settings.
The ResNeXt50~\cite{xie2017aggregated} was used as the backbone with squeeze-and-excitation module and dilated convolutions\rv{~for the video branch. To take images/frames as input for the downstream tasks, following~\cite{jiao2020selfvs}, within an interval of a video clip, we randomly sample a few frames (2 in this work due to the memory limitation) to represent the video clip, and each of them is fed into the encoder for the feature extraction. The corresponding features are concatenated subsequently as the video feature}.
\rv{Following~\cite{jiao2020selfvs}, the audio branch shares the same network architecture as the video branch but is optimised separately and with different inputs. The speech audio was extracted with a 0.6s interval and resampled to 24kHz before being converted to a 2D log-spectrogram representation (of size $256\times256$) using a short-time Fourier transform. Then this log-spectrogram is fed into the audio net for the following representation learning.}
The text branch and audio branch are designed in a similar manner except that we used the word2vec model~\cite{mikolov2013distributed} for text embedding extraction, while the remaining joint modelling with the video embedding is performed in the same way.
We also tried a more advanced text embedding model (BERT~\cite{devlin2018bert}) in replacement of the word2vec model, but did not observe any noticeable improvement. We suspect the reason lies in the relatively simpler context of our text data, while the BERT model was designed to address the ambiguous context situation where the same word can have more than one meaning depending on the context. Most of our words are high-level anatomical (\eg `bones', `heart', `arms', `legs', `blood', `kidneys', and `elbow'), and are adequately covered by a general corpus. BioBERT is an alternative to BERT that has been trained on biomedical corpora. BioBERT has been shown to work well for specialist biomedical terms such as `antimicrobial' and `transcriptional' \cite{lee2020biobert}. However, despite the medical context in which our data is acquired, our vocabulary is more conversational because it is spoken between a clinical professional and the patient. It did not seem necessary to use such a specialised embedding model. In terms of optimisation, we use a similar combined objective as in the aforementioned audio-video modelling, \ie correspondence modelling with contrastive learning.
The weighting parameters in the final loss (equation~(\ref{eq:loss})) are empirically set to be equal after normalising to the same scale. 
\rv{We also did an analysis on the weighting parameters by setting different values to each of the parameters, and found that the model performs the best when they are set to be equal.}
The whole model was trained end-to-end with the stochastic gradient descent (SGD) optimiser and the learning rate was initialised as $10^{-3}$ and decayed by the scale of 10 for every 20 epochs. The batch size was set to 40. The visual signal was scaled to $256\times256$ and centre cropped to $224\times224$ to feed into the GPU. We implemented with the PyTorch deep learning framework on a workstation equipped with NVIDIA GPU cards (Titan V and Tesla V100).

\subsection*{Representation Quality Assessment}

\begin{figure*}
\includegraphics[width=\textwidth]{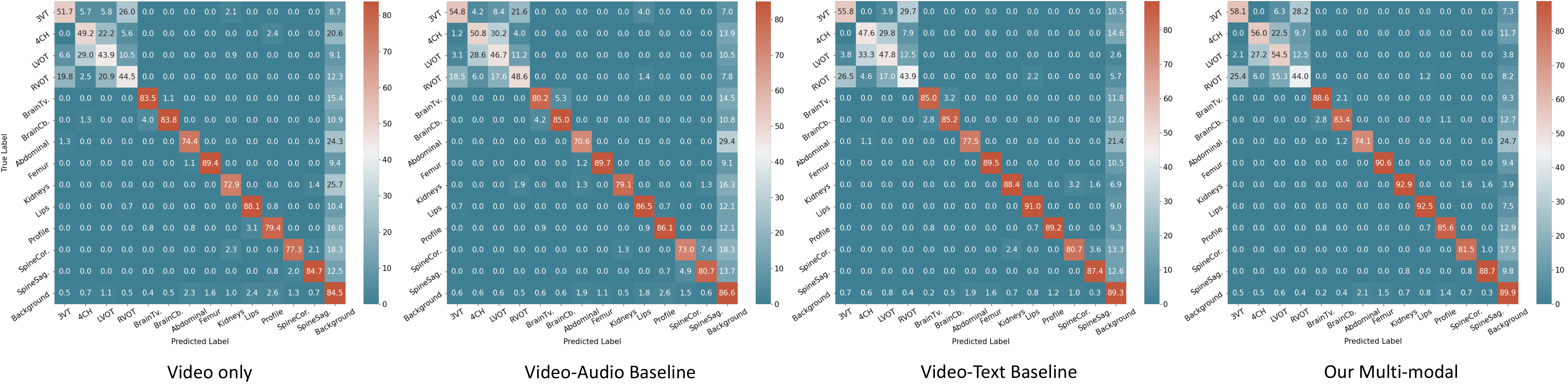}
\caption{\rv{Confusion matrices of the representation learning methods on the SPD task.}}
\label{fig:cms}
\end{figure*}

\paragraph{Standard plane detection}
As mentioned above, the quality of the learned multi-modal representations is assessed by transfer learning to downstream tasks. First we perform evaluation on the standard plane detection (SPD) task. The 14 categories used in this task are:  three-vessel tracheal view of the heart (3VT),  four-chamber view of the heart (4CH), left ventricular outflow tract of the heart (LVOT), right ventricular outflow tract of the heart (RVOT),  transventricular plane of the brain (BrainTv.),  transcerebellar plane of the brain (BrainCb.), abdomen, femur, kidneys, lips, profile, spine in the coronal plane (SpineCor.), spine in the sagittal plane (SpineSag.) and background.
\rv{The performance of this classification task is reported in \cref{tab:spd}, where we include the results of using random initialisation, only video modality, video-audio baseline~\cite{jiao2020selfvs}, video-text baseline and our multi-modal approach (\textit{Ours}) as initialisation for the SPD model. 
Here the \textit{Video} refers to the self-supervised representation learning approach (with a frame order prediction pretext task) using only video data for pre-training; and the \textit{V-A Base} refers to the video-audio self-supervised representation learning approach proposed in~\cite{jiao2020selfvs}.
}
\rv{The \textit{V-T Base} is the baseline using only the video and text branch (without the proposed keywords spotting SIG module) as shown in Figure~\ref{fig:arch}c.}
Additionally, we also report the results of two fully-supervised models which were trained with manual annotated ground-truth labels: \emph{ImageNet Init.} for using the ImageNet~\cite{russakovsky2015imagenet} dataset as training data, and \emph{US-Sup} for using US data with expert annotations as supervision for the same standard plane detection task during pre-training.
From the quantitative results shown in~\cref{tab:spd}, we can see that by using the additional audio data, the performance of SPD can be improved compared to using video only. When using the proposed multi-modal approach, the performance is further improved by a large margin (\eg 72.7$\rightarrow$77.2 for the precision). 
\rv{It can be seen that when only using the video and text branches, the model also performs quite well, suggesting the effectiveness of the data processing and text-related modules as well as the joint modelling.}
It is interesting to see that even though the \emph{ImageNet Initialisation} was trained with human annotations in a fully-supervised manner, the proposed approach still performs much better. As \emph{US-Sup} was pre-trained with the exact same task of SPD and with human annotations, it can be considered as an upper bound.
To better understand how the representation learning methods perform in more detail, we also present the confusion matrix over all the 14 categories, as shown in~\cref{fig:cms}. From the confusion matrix, it can be seen that the cardiac-related categories (\ie \emph{3VT, 4CH, LVOT, RVOT}) are relatively poorly performing; it is well known that these are hard even for a human expert. However, when we compare the \emph{Video-Audio Baseline} with the \emph{Video only}, the performance of these categories is greatly improved, suggesting the effectiveness of the auxiliary audio modality. Building upon that, the proposed multi-modal spatial aware approach further improved almost all the categories significantly. For instance, \emph{LVOT}: 46.7$\rightarrow$54.5 and \emph{Kidneys}: 79.1$\rightarrow$92.9. This fine-grained analysis further validates the effectiveness of the proposed method.

\begin{table*}[t]
  \caption{\rv{Evaluation results on standard plane detection (mean$\pm$std.[\%]). Best performance is marked in \textbf{bold}. Note the \textcolor{gray}{methods} on the right side are fully-supervised using external manual annotations. All the three metrics are the higher the better.}}
	  \centering
	  \begin{tabular}{@{}l|ccccc|cc@{}}
		    \toprule
		    & Rand.Init. & Video & V-A Base & \rv{V-T Base} & Ours & ImageNet Init. & US-Sup \\
		    \midrule
		    Precision & 70.4{\small$\pm$2.3} & 71.9{\small$\pm$2.0} & 72.7{\small$\pm$1.8} & \rv{75.6{\small$\pm$1.4}} & \textbf{77.2}{\small$\pm$1.4} & \textcolor{gray}{74.6{\small$\pm$1.8}} & \textcolor{gray}{87.1{\small$\pm$1.9}} \\
		    Recall & 64.9{\small$\pm$1.6} & 71.7{\small$\pm$3.5} & 73.3{\small$\pm$2.4} & \rv{75.4{\small$\pm$1.6}} & \textbf{77.3}{\small$\pm$1.8} & \textcolor{gray}{71.2{\small$\pm$1.9}} & \textcolor{gray}{81.0{\small$\pm$2.7}} \\
		    F1-score & 67.0{\small$\pm$1.3} & 71.5{\small$\pm$2.4} & 72.6{\small$\pm$1.7} & \rv{75.1{\small$\pm$0.2}} & \textbf{76.8}{\small$\pm$1.0} & \textcolor{gray}{72.5{\small$\pm$1.8}} & \textcolor{gray}{82.7{\small$\pm$1.6}} \\
		    \bottomrule
		  \end{tabular}
  \label{tab:spd}
\end{table*}

\begin{figure*}[h]
\includegraphics[width=\textwidth]{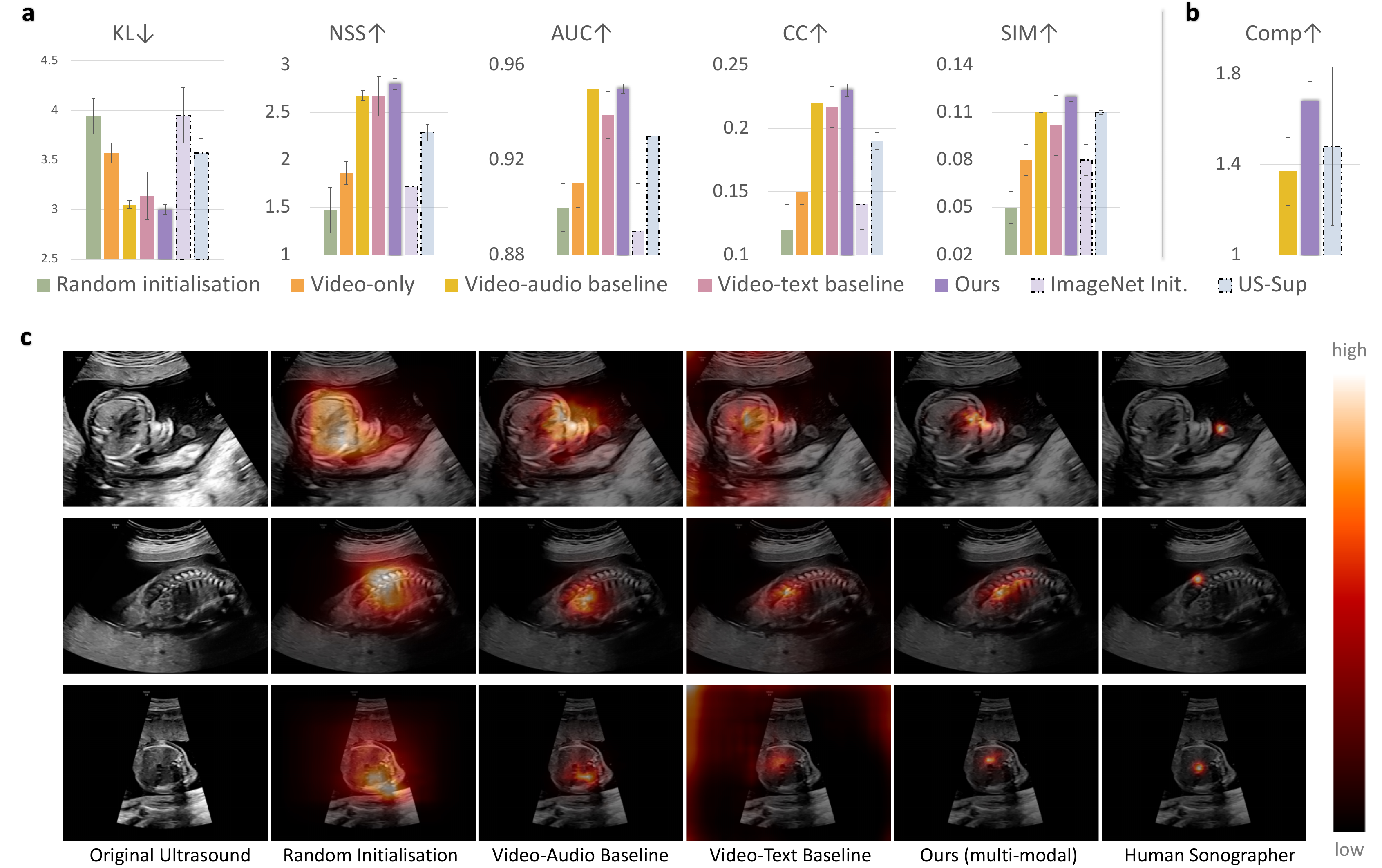}
\caption{\rv{\textbf{Evaluation on eye-gaze saliency prediction.} Quantitative performance (\textbf{a}) evaluated by conventional metrics, and (\textbf{b}) with the newly proposed \textit{Comp} metric. $\downarrow$ indicates the lower the better while $\uparrow$ means the higher the better.} (\textbf{c}) Qualitative performance on eye-gaze saliency prediction.}
\label{fig:salnum}
\end{figure*}

\paragraph{Eye-gaze saliency prediction}
As in our PULSE dataset, the eye-gaze tracking information of the sonographers was captured while performing  US, in addition to the above SPD task; this allows us to take the eye-gaze saliency prediction as an additional downstream task to evaluate the quality of the learned representations. Similar to the SPD task, the pre-trained model was loaded to the saliency prediction model followed by fine-tuning. The only difference is that the last layers were modified so that the model is able to predict a spatial saliency map instead of a single category.
The quantitative performance is reported in \cref{fig:salnum}, with the evaluation metrics mentioned above. Again, we can see that with the addition of the audio modality, the model performs much better when compared to using only video data. 
\rv{The textual information also helps the representation learning, with similar performance as the video-audio baseline and outperforming the single-modal approach. When jointly trained with video, audio and text, the proposed multi-modal approach further boosts the performance by a large margin.}
Different from the results for the standard plane detection task (\cref{tab:spd}), in this regression task we can see that our method outperforms the supervised settings (\emph{ImageNet Init.} and \emph{US-Sup}), suggesting that our approach has better generalisability. We also present qualitative examples in~\cref{fig:salnum}c, in which the predicted saliency map is overlaid to the original US image. This qualitative comparison further validates the effectiveness of the proposed method.
An interesting observation is that, when comparing our approach with the video-audio baseline, we can see (in the qualitative results) that our predicted saliency is more compact and closer to the human performance, though for some of the quantitative metrics shown in~\cref{fig:salnum}a, the difference is not that significant (\emph{Video-audio baseline} vs. \emph{Ours}). As clarified above, this is mainly due to the fact that the metrics in~\cref{fig:salnum}a are proposed for natural images instead of medical imaging. When we conduct a further evaluation by using the newly proposed \emph{Comp} metric (as in~\cref{fig:salnum}b), we can see that the proposed approach has a much better compactness performance, aligning well with the qualitative results.


\subsection*{Ablation Study}

\begin{figure*}
\includegraphics[width=\textwidth]{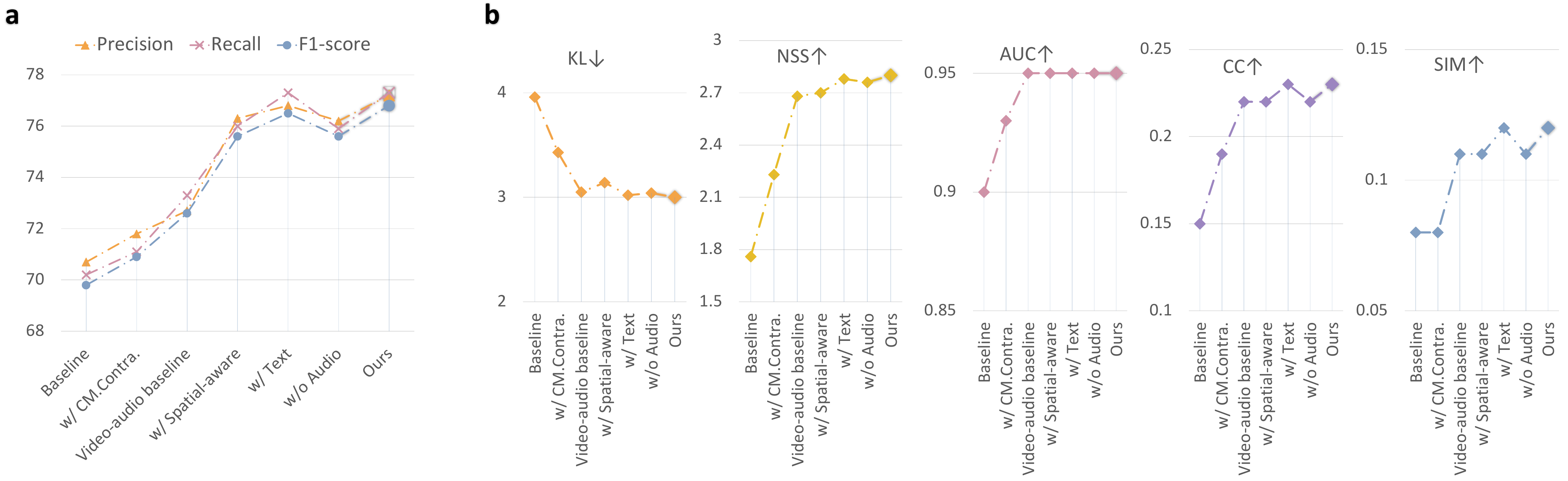}
\caption{\textbf{Ablation study on each of the proposed strategies for two downstream tasks.} (\textbf{a}) Standard plane detection. (\textbf{b}) Eye-gaze saliency prediction, where $\downarrow$ indicates the lower the better while $\uparrow$ means the higher the better.}
\label{fig:abl}
\end{figure*}

In order to analyse the effectiveness of each component in the proposed framework, we conducted a controlled experiment (\ie ablation study), as shown in~\cref{fig:abl}. Performance on the above two downstream tasks is reported with the corresponding evaluation metrics.
Here we chose the baseline as the very basic audio-video correspondence modelling (Figure~\ref{fig:arch}a). Then we compared this to the performance of adding cross-modal contrastive learning (\emph{CM. Contra.}), the aforementioned video-audio baseline, the spatial awareness, the additional textual information, only using visual and text path for training (\emph{w/o Audio}) and the SIG module.
We can see that each component contributes to the performance gain of our proposed approach, validating their effectiveness.

\subsection*{Audio-Guided Visual Anatomy Localisation}

\begin{figure*}[t]
  \includegraphics[width=\textwidth]{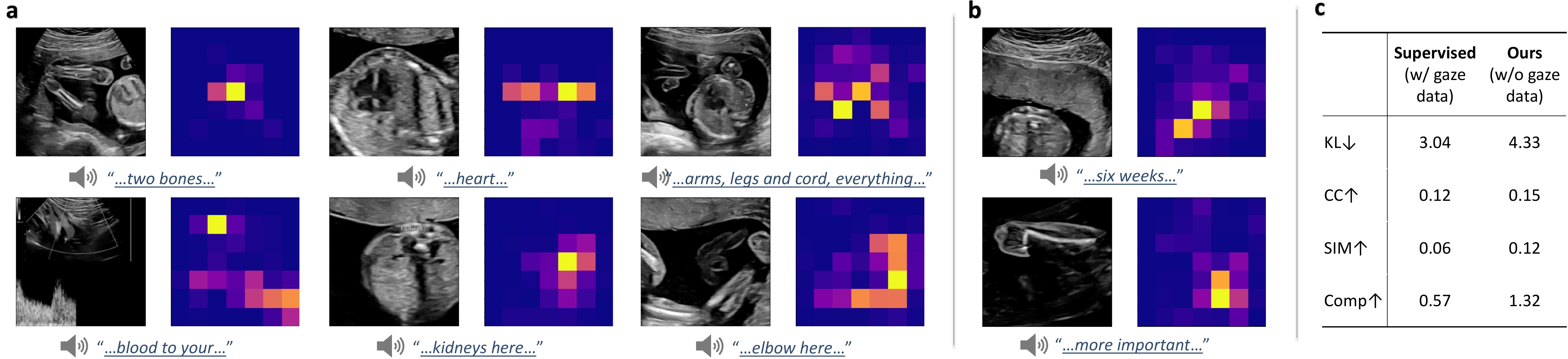}
    \caption{\textbf{Performance on audio-guided visual anatomy localisation.} (\textbf{a}) Performance on normal cases where audio is related to the visual anatomy, while in (\textbf{b}) the audio is unrelated to anatomy. In each sample, the left shows the input US with the corresponding speech audio (transcription) below; the right side shows the estimated localisation map from the proposed method. Higher response is in yellower colour while lower in bluer colour.
    (\textbf{c}) Quantitative evaluation of the audio-guided visual localisation task.}
    \label{fig:local}
\end{figure*}

We have shown that the proposed multi-modal audio-visual modelling framework is able to perform audio-guided visual anatomy localisation with no extra human interventions. In this section, we showcase how this can be achieved.
During the training of the multi-modal representation learning framework, audio (as well as derived text) embedding perceive the spatial information of the visual data well, by the fusion module. As a result, after the training, given input audio the model is able to highlight the high-response visual regions and such regions are usually related to anatomies due to the training target.
Note that instead of visualising landmarks, our model shows the association between the visual signal and the audio input, trying to highlight the corresponding anatomical structures.
In Figure~\ref{fig:local}a, we showcase examples of this audio-guided visual anatomy localisation task. We can see that our model is able to correctly localise the anatomies that the speech audio contains, \eg \emph{heart, kidneys} \etc.
This demonstrates the quality of the learned joint multi-modal representations and also illustrates a useful application towards anatomy localisation.
\rv{The strategy for the audio-visual anatomy localisation task is the same setting as in our work mentioned above, \ie~the speech audio data is fed into the pretrained audio net together with the video data captured from the ultrasound machine being fed into the video net. Then the model is able to highlight the high-response visual anatomical regions accordingly. Under our experimental settings, the time to process a $244\times244$ frame and generate the corresponding response map is $0.08$s, \ie over 12 FPS (frames per second), which is acceptable for applications.}
\rv{As mentioned above, it is infeasible to acquire accurate clinical annotations for this task and we instead seek help from the eye-gaze saliency data as a reference to provide an approximate quantitative evaluation here. Specifically, we compare the localisation map of the prediction from our model and the eye-gaze saliency from the captured gaze data, and measure their similarity on a separate unseen test set with both audio and gaze data. The results are shown in \cref{fig:local}c, using the evaluation metrics of KL, CC, SIM and Comp (NSS and AUC are discarded as in this task we are not measuring human fixations). For comparison, we also include a saliency prediction method that was trained in a fully-supervised manner with ground-truth supervision from eye-gaze saliency data, and applied to the same test data. Note our method in this audio-guided visual localisation task never saw any gaze/saliency data but was only guided by the audio signal. From the results we can see that even without seeing any eye gaze data, our method still performs well and better (3 out of 4 metrics) than the solution that was trained with eye-gaze data as ground-truth supervision.}
As described above, the content of the speech audio is not always related to anatomical regions. In the case where the audio does not relate to any anatomy, the model may fail to give meaningful localisation. Examples of such ``failure cases'' are shown in Figure~\ref{fig:local}b for reference. Interestingly, though the audio content does not relate to any anatomy, the model still tries to reason about some localisation, \eg the model is looking at the anatomy and its neighbours to try to `figure out' the ``\underline{six weeks}'' narration.

\section*{Discussion}

\paragraph{Limitations}
Although this work made the first attempt towards learning multi-modal representations in real-world clinical settings and demonstrated meaningful representations, there are some limitations that may prevent the proposed approach from being applied more generally. In this work, we only validated the proposed model on clinical fetal US data. The proposed framework is general and there are no assumptions which make it obstetrics specific, however it has not been validated in other clinical settings as yet to prove generalisability. Another limitation is that the current model is designed to localise \emph{anatomy-related} regions and not more general concepts (as shown in~\cref{fig:local}b). The main reason is that the most consistent audio-video pairs are related to anatomy in our case; if other well-paired training data for other concepts were available, this limitation could be addressed accordingly.

\paragraph{Potential impact and benefits}
The study was conducted using data from one hospital site in Oxford, thus the corresponding conclusions may not identical for data from other sites. However, data was collected in a real-world setting; this also explains the presence of audio signals that are not relevant to anatomical regions, for example conversations between the sonographer and prospective parents. Thus, we believe that the model should apply to most similar settings, though including more data from other clinical environments could strengthen the model.
There are several strengths of this work. We have made the first attempt towards multi-modal clinical representation learning with a particular focus on video-audio data modalities. This enables us to have a better and more thorough understanding of the clinical data and eliminates the requirement of human expert annotations. 
{In some clinical settings, including the UK where this study was conducted, sonographers will communicate their findings and provisional information to the patient, thereby providing an additional modality to the data science of sonography. And sonographers usually impart such information during the scan, and confirming the normality of views or absence of abnormalities.}
The proposed model can also be utilised to help localise anatomical regions of interest, by only using speech audio, \ie \emph{talk to the screen, and show from tell}. This has both scientific and clinical contributions. Data annotation can be made much easier by only verbally describing the regions that need to be labelled. The proposed model may be useful to help the training of new sonographers.

\section*{Conclusion}
In this work, we presented a study on multi-modal representation learning in clinical settings, with a particular focus on audio-visual data. A new multi-modal framework has been proposed with additional textual data and a selective gating module. Extensive experimental analysis on a large-scale clinical dataset shows that the proposed approach can learn meaningful representations without additional human expert intervention. The learned representations were shown to be effective in several downstream clinical tasks. We also introduced a new clinical application, audio-guided visual localisation, which may be useful to both clinicians and patients in video interpretation.

\section*{Acknowledgements}
The authors would like to thank the support from the EPSRC Programme Grant Visual AI EP/T028572/1, the ERC Project PULSE ERC-ADG-2015 694581, and the support of NVIDIA Corporation with the donation of the GPU. A.T.P. is supported by the Oxford Partnership Comprehensive Biomedical Research Centre with funding from the NIHR Biomedical Research Centre (BRC) funding scheme.

\bibliographystyle{ieee_fullname.bst}
\bibliography{egbib}

\end{document}